\documentclass[conference]{IEEEtran}
\IEEEoverridecommandlockouts
\usepackage{cite}
\usepackage{amsmath,amssymb,amsfonts}
\usepackage{algorithmic}
\usepackage{graphicx}
\usepackage{textcomp}
\usepackage{xcolor}
\usepackage{amsthm} 

\def\BibTeX{{\rm B\kern-.05em{\sc i\kern-.025em b}\kern-.08em
    T\kern-.1667em\lower.7ex\hbox{E}\kern-.125emX}}
\begin{document}

\title{Feature Learning and Classification in Neuroimaging: Predicting Cognitive Impairment from Magnetic Resonance Imaging\\
{\footnotesize 
}
}

\author{\IEEEauthorblockN{1\textsuperscript{st} Shan Shi}
\IEEEauthorblockA{\textit{Department of Mathematics and Statistics} \\
\textit{University of Victoria}\\
Victoria, Canada \\
sshi@uvic.ca}
\and
\IEEEauthorblockN{2\textsuperscript{nd} Farouk Nathoo}
\IEEEauthorblockA{\textit{Department of Mathematics and Statistics} \\
\textit{University of Victoria}\\
Victoria, Canada \\
nathoo@uvic.ca}}

\maketitle

\begin{abstract}
Due to the rapid innovation of technology and the desire to find and employ biomarkers for neurodegenerative disease, high-dimensional data classification problems are routinely encountered in neuroimaging studies. To avoid over-fitting and to explore relationships between disease and potential biomarkers, feature learning and selection plays an important role in classifier construction and is an important area in machine learning. In this article, we review several important feature learning and selection techniques including lasso-based methods, PCA, the two-sample t-test, and stacked auto-encoders. We compare these approaches using a numerical study involving the prediction of Alzheimer's disease from Magnetic Resonance Imaging.
\end{abstract}


\section{Introduction}
Due to the rapid innovation of technology, collection of a massive amount of high-dimensional data is becoming increasingly easier and inexpensive. For example, in studies of Alzheimer's disease (AD), neuroscientists can collect large datasets representing magnetic resonance images (MRI) and functional MRI images for each subject in a group of tens to hundreds. When these data are used for classification, the large number of potential features presents a huge challenge to classical classification methods. When the dimensionality is high compared to the sample size, classical methods may not perform well or may even break down. This well-known phenomenon is referred to as the {\em{curse of dimensionality}}. Taking classification with high-dimensional genetic data for example, the problem can be difficult because of the existence of many noise features (e.g. SNPs that not related to the phenotype of interest), which don't contribute to classification accuracy but do contribute noise that can weaken the discriminative power of the classifier. Reference \cite{f4} provides detail about the impact of high-dimensionality on classification when the dimension $p$ diverges as the sample size $n$ increases. 

When dealing with high-dimensional classification problems, dimension reduction or feature selection methods are widely employed to facilitate classifier construction. In the context of machine learning, feature learning refers to the process of engineering features from raw data, while feature selection is a special case of feature learning that involves selecting the most important, in some sense, features out of a larger collection. Feature selection can help to (i) understand the underlying relationships by removing irrelevant features; (ii) overcome the noise accumulation effect \cite{f7}; (iii) take advantage of prior knowledge.

Feature selection methods can be divided into three subcategories, namely, filters, wrappers and embedded methods. Filters represent methods which select features before applying a machine learning algorithm for classification. Wrappers evaluate the performance of a specified classifier on selected features and keep adding to or removing from the subset until some criterion is met. Embedded methods are used to incorporate the feature selection into the process of training a classifier.

Dimension reduction techniques such as principal-component analysis (PCA), create new features that are a function of the original features - linear combinations in the case of PCA. Moving away from linear features, an auto-encoder is an alternative dimension reduction approach used to obtain features that are nonlinear functions of the original input data. One drawback of such methods is that the new features are difficult to interpret, since they are obtained by using all of the original features and a nonlinear transformation. Nevertheless, dimension reduction methods are powerful and effective tools that can often capture most of the variability of the original dataset. 

In this paper, we provide a review of four popular feature learning methods for high-dimensional data and make comparisons within the context of classification of disease from neuroimaging data. More specifically, we consider the problem of predicting Alzheimer's disease from Magnetic Resonance Imaging (MRI) data. In Section II we provide a review and general discussion for each of the four methods considered in our study. In Section III we apply and compare these methods to the problem of classifying Alzheimer's disease using MRI data. We conclude with a discussion in Section IV.


\section{Feature Learning}

For simplicity, we will only consider the problem of binary classification with high-dimensional data. Specifically, given n data points, $\{\left(X_{i},Y_{i}\right), i=1\dots, n\}$ , sampled from a joint distribution of $\left(X,Y\right)\in \mathbb{R}^{p}\times \{0,1\}$. There are $n_{0}$ sample points from class $Y=0$ and $n_{1}$ from class $Y={1}$ .The goal is to predict $Y$ given a new $X$. 



\subsection{The Lasso}

Given a data matrix $\mathbf{X}\in \mathbb{R}^{n\times p}$, we here assume that each column  of $\mathbf{X}$ is standardized. The lasso proposed by Tibshirani in 1996 and the lasso estimate $\hat{\beta}$ in the linear model is:
\[
\hat{\beta}(\lambda)=\underset{\beta}{\mathrm{argmin}}\left(\|Y-\mathbf{X}\beta\|^{2}_{2}/n + \lambda \|\beta\|_{1}\right),
\]
where $\|Y-\mathbf{X}\beta\|^{2}_{2}= \sum_{i=1}^{n}\left(Y_{i}-X_{i}^{T}\beta\right)^{2}$, $\|\beta\|_{1}=\sum_{j=1}^{p}|\beta_{j}|$ and the penalty parameter $\lambda>0$. An important property of the lasso is that the estimator is sparse so that $\hat{\beta}(\lambda)_{j}=0$ for some $j$'s. As a result, the features with $\hat{\beta}(\lambda)_{j}\neq 0$ can be thought of as those features that have been selected by the lasso. Applications of lasso-based feature learning in neuroimaging classification are considered in \cite{s1}. They propose to fit a linear regression model with the lasso to the training data for feature selection. Then a classifier is trained using the selected features.

It is worth mentioning that lasso-based feature selection methods are easy to implement and could help improve classification accuracy but the selected features such as genes or SNPs may not be the right ones for understanding genome-wide disease association, due to incidental endogeneity \cite{f5}. Generally speaking, the more covariates that are collected which are thought to potentially related to the response, the less likely all covariates are uncorrelated with the residual noise. In other words, some of the covariates are very likely incidentally correlated with the residual noise. When this fundamental assumption $\mathbb{E}(\epsilon X)=0$ cannot be guaranteed to hold, \cite{f6} show that this causes model selection inconsistency of the penalized least-squares method. To overcome incidental endogeneity, different methods have been proposed. Reference \cite{f6} proposes a penalized method, {\em focused generalized method of moments} (FGMM), based on the idea of {\em over identification}; \cite{r1} suggest a novel way of dealing with this problem, relying on sample splitting to perform valid inference after model selection.

\subsection{Two-Sample t-Test}
The two-sample t-test is an easily employed and simple feature selection method when dealing with classification and a large number of features. Reference \cite{f1} show that under some mild conditions, the t-test can correctly select all important features with high probability. 
Let $\bar{X}_{kj}=\sum_{Y_{i}=k}X_{ij}/n_{k}$ and $S^{2}_{k}=\sum_{Y_{i}=k}(X_{ij}-\bar{X}_{kj})^{2}/(n_{k}-1)$ be the sample mean and variance of the $j$-th feature in class $k$, where $k=0,1$, and $j=1,\dots,p$. The two-sample t-test statistic for feature $j$ is defined as: $$T_{j}=\frac{\bar{X}_{0j}-\bar{X}_{1j}}{\sqrt{S^{2}_{0j}/n_{0}+S^{2}_{1j}/n_{1}}},\text{ } j=1,\dots,p.$$ Assume that $\mu=\mathbb{E}(X|Y=0)-\mathbb{E}(X|Y=1)=(\mu_{1},\dots,\mu_{p})$ is sparse with only the first $s$ entries nonzero and some other mild conditions, the following result holds: $$
\mathbb{P}\left(\underset{j\leq s}{\mathrm{min}}|T_{j}|> x, \underset{j>s}{\mathrm{max}}|T_{j}|<x\right)\rightarrow 1,\text{ as }n,p\rightarrow\infty,$$
where $x$ is some positive constant. The above result tells us that, with high probability, all important features can be selected according to the two-sample t-test if an appropriate threshold value $x$ is chosen. In practice, it is hard to choose a good threshold value to find all important features. Reference \cite{f1} points out that it is not necessary to use all relevant features because of the possible existence of many faint features, which yield weak signals. When using the independence classification rule \cite{f1}, the optimal number of features, $m$, these authors suggest can be estimated by minimizing the following misclassification rate upper bound:
$$\hat{m}_{opt}=\underset{1\leq m\leq p}{\mathrm{argmax}}\frac{1}{\hat{\lambda}_{max}^{m}}\frac{n\left[\sum_{j=1}^{m}T^{2}_{(j)}+m(n_{0}-n_{1})/n\right]^{2}}{mn_{0}n_{1}+n_{0}n_{1}\sum_{j=1}^{m}T^{2}_{(j)}},$$ where  $T^{2}_{(1)}\geq T^{2}_{(2)}\geq \cdots \geq T^{2}_{(p)} $ are the ordered squared t-test statistics, $n=n_{0}+n_{1}$, and $\hat{\lambda}_{max}^{m}$ is the largest eigenvalue of the sample correlation matrix $\mathbf{R}^{m}$ of the truncated observation, that is, only the first $m$ highest t-statistic features are considered. For any classifier other than the independence classification rule, $m$ should be treated as a tuning parameter and cross-validation can be used to find the optimal $m$.

\subsection{Principal Component Analysis (PCA)}
 
Given a data matrix $\mathbf{X}\in \mathbb{R}^{n\times p}$, which in the current context represents $n$ observations on a set of $p$ features, we have $n$ data points $\{X_{i},i=1,\dots,n\}$ in $\mathbb{R}^{p}$ with $\mu_{0}=\mathbb{E}(X|Y=0)$ and $\mu_{1}=\mathbb{E}(X|Y=1)$. Here, we assume that the variance of each feature is one. The sample covariance matrix is $\mathbf{S}=\frac{1}{n}\sum_{i=1}^{n} \left(X_{i}-\bar{X}\right)\left(X_{i}-\bar{X}\right)^{T}$, where $\bar{X}=\frac{1}{n}\sum_{i=1}^{n}X_{i}$. 

Principal components (PCs) can be derived and viewed from two different but equivalent ways. We refer readers to \cite{j2} for technical details and also to \cite{j3} for a recent review. The first method is by finding the directions of maximum variance. 
The first $r$ principal component directions, $\mathbf{V}_{p\times r}$, can be found by solving the following optimization problem, where $r\leq p$ $$\mathbf{V}_{p\times r}=\underset{\mathbf{A}:\mathbf{A}^{T}\mathbf{A}=\mathbf{I}_{r}}{\mathrm{argmax}}trace\left(\mathbf{A}^{T}\mathbf{S}\mathbf{A}\right).$$

The second approach to finding the principal components is based on minimizing what is referred to as the reconstruction error, where we seek the matrix $\mathbf{A}_{p\times r}$ with orthonormal columns i.e. $\mathbf{A}^{T}\mathbf{A}=\mathbf{I}_{r}$ which minimizes:
$$
\frac{1}{n}\sum_{i=1}^{n}\|\left(X_{i}-\bar{X}\right)-\mathbf{A}\mathbf{A}^{T}\left(X_{i}-\bar{X}\right)\|^{2}_{2}.$$ The projection matrix $\mathbf{H}=\mathbf{A}\mathbf{A}^{T}$ projects each sample point $X_{i}$ to the $r$ dimension subspace spanned by the columns of $\mathbf{A}$. The easiest way of applying PCA for high-dimensional classification is to reduce the dimensionality of the data by replacing $X\in \mathbb{R}^{p}$ by its first $m<p$ principal components. Interestingly, \cite{j2} points out that the PCs with high variance are not necessarily useful for separation between two populations. This is because the direction of highest variance may not well-separate the sample means after projection. 
\cite{j2} suggests using the Mahalanobis distance between two populations defined by a subset of PCs to measure the discriminatory power of that subset. Specifically, if the $k$th PC with direction vector $v_{k}$ maximizes the quantity $$\theta_{k}=\frac{[{v^{T}_{k}(\bar{X}_{0}-\bar{X}_{1})}]^{2}}{l_{k}},$$ where $l_{k}$ is the sample variance of the $k$th PC, the $k$th PC has the largest discriminatory power. Intuitively speaking, what the author suggests is that we have to use the information provided by the response variable $Y$ to guide the algorithm. 

In the case where $p\gg n$, ordinary PCA may perform poorly as discussed in \cite{j4}. This poor performance occurs because the eigenvectors of the sample covariance matrix are not necessarily close to the true eigenvectors, in which case the first few principle components cannot capture most of the information contained in the data. 


\subsection{Stacked Auto-encoder}
An auto-encoder is a feature learning approach that can be thought of as a generalization of PCA. While PCA is based on projecting the data onto linear subspaces, auto-encoders enable us to deal with a curved manifold in the input space, so the data are represented by projections on the curved manifold. 
Let $f_{\theta}$ and $g_{\theta}$ be functions which we will call the the encoder and decoder respectively. For each data point $X_{i}\in \mathbb{R}^{p}$, we compute a {\em{representation}} $h_{i}$ from $X_{i}$ using the encoder as follows: $h_{i}=f_{\theta}(X_{i}).$ The decoder $g_{\theta}$ is then a mapping from the feature space back to the input space and is intended to produce a reconstruction of the input data $\hat{X}_{i}=g_{\theta}(h_{i}).$ Like PCA, the parameters $\theta$ characterizing the encoder and decoder are estimated by minimizing the reconstruction error: $$L(\theta)=\sum_{i=1}^{n}\|X_{i}-g_{\theta}(f_{\theta}(X_{i}))\|^{2}_{2}.$$ If the input domain is unbounded, the most common choices for the encoder and decoder are: 
\begin{align*}
f_{\theta}&=\sigma(b+\mathbf{W}X_{i})\\
g_{\theta}&=d+\mathbf{W}^{T}h_{i}
\end{align*}
where $\sigma(x)$ is the sigmoid function, $b$, and $d$ are called bias vectors and $\mathbf{W}$ is the weight matrix. If the identity function is used instead of the sigmoid function, the resulting auto-encoder will learn the subspace equivalent to that learned by PCA \cite{b4}. 

An auto-encoder can be used as a a building block in a more complex structure known as a stacked auto-encoder (SAE) \cite{b3}.  This structure is based on a sequence of auto-encoders that are stacked one on top of the other. The input $X_{i}$ is encoded sequentially by each encoder. Specifically, the first auto-encoder yields the representation $h^{(1)}_{i}=f^{(1)}_{\theta}(X_{i})$, then the output $h^{(1)}_{i}$ is the input of the second auto-encoder to learn the second representation $h^{(2)}_{i}$ and so on. 

One of the most important characteristics of SAEs arises when they are used to form deep architectures with several layers which can capture highly non-linear functions of the raw data. Thus, deep architectures can lead to abstract representations. In order to train SAEs, one typically trains the first auto-encoder using the training data as input. Given the result of the first auto-encoder, the representation $\{h^{(1)}_{i}\}$ is used as input to train a second auto-encoder, and this process continues until the last auto-encoder. Finally, at the top of the SAE, a softmax classifier layer is added to do classification with the learned representations. Finally, after the sequential stage-wise training is complete, the entire SAE and the softmax layer is trained jointly as one feed-forward neural network (SAE-classifier) to improve all of the weights in the combined SAE and classifier.  The loss function used to train such SAE-classifers is cross entropy loss (the log-likelihood) with L$2$-regularization \cite{g1}. Such a combined strategy is often called fine-tuning. Aside from being used to create feature-vectors for a classifier, the outputs of an SAE can be fed into a classifier as extra inputs in addition to the original raw inputs\cite{s1}. 
\section{Experimental Evaluation}
In this section, we compare the effectiveness of different feature learning methods by considering a binary classification problem where the goal is to use MRI data and classify subjects into one of two groups: those with Alzheimer's disease (AD) or normal controls (NC). For all methods considered, a linear support vector machine (SVM) is employed as the classifier at the final level. 

\subsection{Dataset}

The data are from the Alzheimer's Disease Neuroimaging Initiative ADNI-I study (http://adni.loni.usc.edu). The ages of the subjects comprising the dataset we consider range from 55 to 90 years old. Specifically, in our study, we use 632 subjects. Among them, 144 were diagnosed with AD, 179 are NC, and the remaining 309 were categorized as having mild cognitive impairment (MCI), which in our study are considered as target-unrelated subjects. The MRI data were preprocessed as described in \cite{z1} and the preprocessed images are divided into 56 pre-defined regions-of-interest (ROIs) that are known to be related to AD and the gray matter volume for each ROI is computed. The specific ROIs used in our study are outlined in \cite{g2} and the result is a vector of 56 MRI-derived features.

\subsection{Experimental Setup}
We randomly split the dataset into a training set and a test set. The size of the test set is $20\%$ of the total dataset. We standardize each feature to have zero mean and unit variance in the training set. We use 10-fold cross-validation on the training set to choose the hyperparameters for each method, for example the number of PCs, and tuning parameters associated with penalty terms. We use the test set to evaluate the classification accuracy. This procedure is repeated 100 times to achieve a better evaluation of the classification accuracy. For simplicity of presentation we let LLF, SAEF and LLF+SAEF denote, respectively, the original low-level features (raw 56-dimensional data vectors), the SAE-learned features, and the combination of LLF and SAEF. 

\subsection{Determination of the Structure of the SAE Model}

Due to the small sample size, we use the whole dataset to choose the structure of the SAE. We first randomly split the data into training and test sets as described above. To find the best structure of the SAE, we train 4 different SAE-classifiers, namely, 40-20-15 (each number corresponds to a hidden layer and represents the number of hidden units in that layer from the bottom to the top of the SAE), 50-25-10, 50-20, and 40-15, on the training data and evaluate the prediction power on the test set. The structure with the highest classification accuracy is selected. For each of the SAE-classifiers, the learning rate and number of iterations used in the gradient descent optimization method are also fixed in this stage \cite{g1}. To avoid over-fitting, the number of iterations is set to be $150$ and learning rate is $0.01$. The optimal structure of the SAE model using this approach is found to be 40-15.  

\subsection{Semi-Supervised Approach}

In addition to the supervised approach, the semi-supervised approach is also considered when training the SAE-classifier. That is, we first use both the target-related training and target-unrelated samples to train the auto-encoders before fine-tuning. We then use only the target-related training samples to fine-tune the model. We will let semi-SAEF denote this approach. 

\subsection{Feature Learning Methods}
The feature learning methods that we evaluate are as follows:
\begin{itemize}
	\item No feature selection (No FS), LLF, SAEF, semi-SAEF, LLF + SAEF and LLF + semi-SAEF
	\item PCA, where we represent LLF by the PCs with high variance
	\item Two sample t-test to select a subset of LLF
	\item LASSO-based sparse feature learning applied to LLF, LLF + SAEF, LLF + semi-SAEF
\end{itemize}

\subsection{Experimental Results}

Table 1 shows the mean accuracy of the feature learning methods considered. Examining this table, We find that semi-SAEF and SAEF exhibit the best performance, $86.9\%$(SAEF) and $87.7\%$. Directly concatenating the high-level features (SAEF/semi-SAEF) with LLF does not improve the accuracy, e.g. $83.5\%$ (LLF+SAEF) and $84.1\%$ (LLF+semi-SAEF) vs. $83.9\%$ (LLF). The combination of LLF+SAEF/LLF+semi+SAEF and the lasso-based method to select a subset of the features seems to improve the performance slightly, but the performance is no better than LLF+LASSO, e.g. $84.6\%$ (LLF+SAEF+LASSO) and $85.3\%$ (LLF+semi-SAEF+LASSO) vs. $85.7\%$ (LLF+LASSO). Interestingly, the simple approaches LLF+PCA and LLF+t-test perform essentially equivalently to LLF+LASSO. \emph{Overall, the use of a 40-15 stacked auto-encoder in combination with a semi-supervised learning approach yields the highest accuracy in predicting Alzheimer's Disease from MRI.}
\begin{table}[htbp]
	\caption{}{Performance comparison of different feature learning methods in the classification of AD versus NC (in \% accuracy).}
	\begin{center}
		\begin{tabular}{c|ccccc}
			\textbf{} & {LLF}& {LLF+SAEF}& {LLF+semi-SAEF}&{SAEF}&{semi-SAEF} \\
			\hline
			No FS &83.9 & 83.5& 84.1 & 86.9&87.7\\
			Lasso &85.7 &84.6 & 85.3 & &\\
			t-test &85.9 & &  & &\\
			PCA &85.7 & &  & &\\
		\end{tabular}
		\label{tab1}
	\end{center}
\end{table}

\section{Disscussion}
\subsection{Other forms of auto-encoders}
In this article, we have only considered undercomplete auto-encoders, that is, auto-encoders where the representation dimension is less than the input dimension. From our study, it is clear that undercomplete auto-encoders help to capture the most representative features to achieve the goal of dimension reduction. Recently, overcomplete auto-encoders have also been considered and applied successfully to problems like the one we considered here \cite{b4}. In the overcomplete case, the {representation} dimension is greater than the input dimension. In such cases there is a danger that the auto-encoder will simply copy the input to the output without learning any useful features from the data. To prevent this from occurring, regularized auto-encoders can be used and these provide the ability to choose any representation dimension. Aside from the usual approach of regularizing by adding penalty terms to the corresponding objective function, there are more recent and interesting forms of regularized auto-encoders, such as sparse auto-encoders \cite{b5}, denoising auto-encoders \cite{v1} and contrastive auto-encoders \cite{r2}. These alternative auto-encoders will be studied and new results obtained in a follow-up paper. We refer readers to \cite{g1} for more detail.

\subsection{Stability} 
While not considered in the current paper, the stability of a feature selection method is also an important issue that plays a crucial role in biomedical classification problems such as the one we considered here. In a follow-up paper, we will investigate the stability of the feature selection approaches considered here to small perturbations of the data such as the addition or removal of sample points. It is of interest to determine which of the methods considered here will perform the best with respect to stability. This issue is of great importance and a thorough discussion is provided in Yu \cite{y1}.    

\section{Acknowledgment}
This research is supported by an NSERC discovery grant, a Tier 2 Canada Research Chair, and a CANSSI Collaborative Research Team grant to Farouk Nathoo. Data collection and sharing for this project was funded by the Alzheimer's Disease Neuroimaging Initiative (ADNI) (National Institutes of Health Grant U01 AG024904) and DOD ADNI (Department of Defense award number W81XWH-12-2-0012).

\end{document}